%% file: aacl.tex
\pdfoutput=1

\documentclass[11pt]{article}

\usepackage[final]{aacl} 

\usepackage{times}
\usepackage{latexsym}
\usepackage{graphicx}
\usepackage{multirow}
\usepackage{upgreek}
\usepackage{booktabs}
\usepackage{xcolor}
\usepackage{soul} 

\usepackage[T1]{fontenc}

\usepackage[utf8]{inputenc}

\usepackage{microtype}

\usepackage{inconsolata}
\usepackage{xspace}

\newcommand{\dataset}{\textsc{CovidNews-NER}\xspace}

\newcommand{\model}{\textsc{Controster}\xspace}

\usepackage{amsmath}
\usepackage{amsfonts}
\usepackage{amssymb}

\usepackage{cleveref}
\crefformat{section}{\S#2#1#3}
\crefformat{subsection}{\S#2#1#3}
\crefformat{subsubsection}{\S#2#1#3}
\crefrangeformat{section}{\S\S#3#1#4 to~#5#2#6}
\crefmultiformat{section}{\S\S#2#1#3}{ and~#2#1#3}{, #2#1#3}{ and~#2#1#3}
\usepackage{refstyle}
\Crefformat{figure}{#2Fig.~#1#3}
\Crefmultiformat{figure}{Figs.~#2#1#3}{ and~#2#1#3}{, #2#1#3}{ and~#2#1#3}
\Crefformat{table}{#2Tab.~#1#3}
\Crefmultiformat{table}{Tabs.~#2#1#3}{ and~#2#1#3}{, #2#1#3}{ and~#2#1#3}
\Crefformat{appendix}{Appx.~\S#2#1#3}
\crefformat{algorithm}{Alg.~#2#1#3}

\title{How to tackle an emerging topic? \\ Combining strong and weak labels for Covid news NER}

\author{Aleksander Ficek \\
  University of Waterloo \\
  \texttt{acficek@uwaterloo.ca} \\\And
  Fangyu Liu \\
  University of Cambridge \\
  \texttt{fl399@cam.ac.uk} \\\And
  Nigel Collier \\
  University of Cambridge \\
  \texttt{nhc30@cam.ac.uk}
  }
  
\begin{document}
\maketitle
\input{AACL2022/000_abstract}
\input{AACL2022/001_intro}

\input{AACL2022/002_dataset}
\input{AACL2022/003_model}

\input{AACL2022/004_exp}

\input{AACL2022/005_discussion}
\input{AACL2022/006_conclusion}
\input{AACL2022/007_acknowledgments}


\bibliography{aacl.bbl}
\bibliographystyle{AACL2022/acl_natbib}

\input{AACL2022/008_appendix}

\end{document}

%% file: AACL2022/000_abstract.tex
\begin{abstract}

Being able to train Named Entity Recognition (NER) models for emerging topics is crucial for many real-world applications especially in the 
medical domain where new topics are continuously evolving out of the scope of existing models and datasets. 
For a realistic evaluation setup, we introduce a novel COVID-19 news NER dataset (\dataset) and release 3000 entries of hand annotated strongly labelled sentences and 13000 auto-generated weakly labelled sentences. Besides the dataset, we propose \model, a recipe to strategically combine weak and strong labels in improving NER in an emerging topic through transfer learning. 
We show the effectiveness of \model on \dataset while providing analysis on combining weak and strong labels for training. Our key findings are: (1) Using weak data to formulate an initial backbone before tuning on strong data outperforms methods trained on only strong or weak data. (2) A combination of out-of-domain and in-domain weak label training is crucial and can overcome saturation when being training on weak labels from a single source.\footnote{Dataset and code is available at \url{https://github.com/aleksficek/covidnews-ner}.}


\end{abstract}

%% file: AACL2022/001_intro.tex
\section{Introduction}\label{sec:intro}
\vspace{-0.2em}
Named Entity Recognition (NER) is an 
NLP task that involves identifying key entities in text such as person, location, time or organisation. Research around NER has grown rapidly with the adoption of deep learning techniques and has been an integral step to many NLP pipelines \citep{sun2018overview} such as information retrieval, knowledge base completion, and question answering. As NER models have matured to involve deep Transformer \citep{vaswani2017attention} models and achieve greater performance, the demand for more human labelled strong data has followed. This has become a common bottleneck as attaining more strongly labelled data is expensive and time consuming.

\begin{figure}
    \centering
    \includegraphics[width=0.96\linewidth]{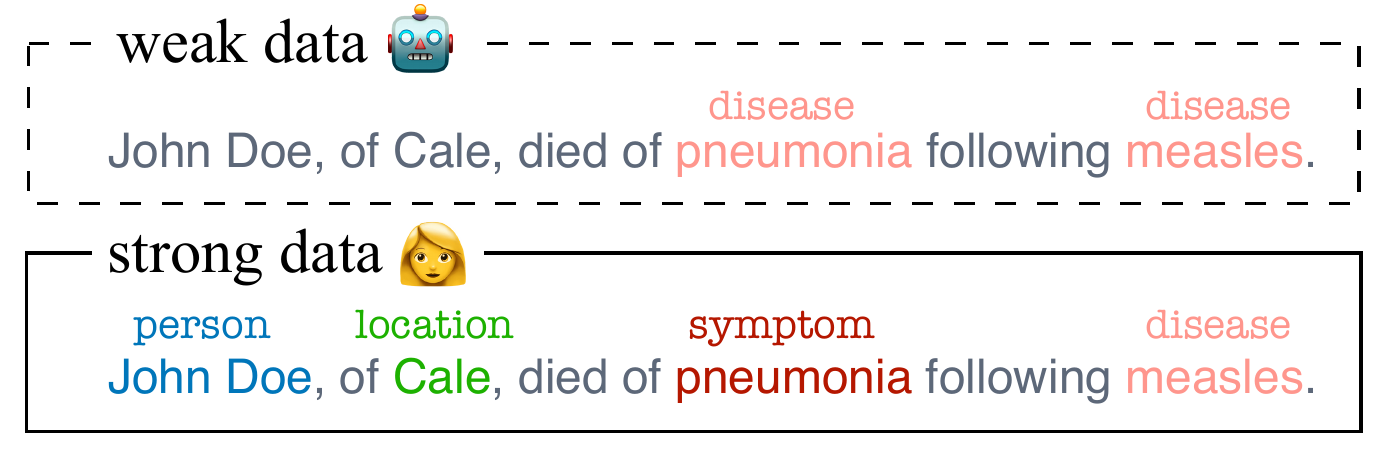}
    \caption{We consider a realistic setup, combining machine-generated noisy weak labels and a small amount of human-generated strong labels for tackling NER in an emerging domain. \scriptsize{(Real name replaced with a fictitious name: \url{https://en.wikipedia.org/wiki/John_Doe}.)}}
    \label{fig:main_fig}
    \vspace{-1.5em}
\end{figure}

To work around the limited amount of strongly labelled data, many have experimented with using lower quality weak data generated by weakly supervised methods. Popular techniques to generate weak data include using knowledge bases and heuristic rule based methods while leveraging multiple sources \citep{multilabel, multibert, labeltool}. All techniques can be applied to any suitable text allowing the methods to generate weak data for any topic. Generating weak labels is especially promising for the medical domain where labelling may require experts to accurately label text and common vocabulary is constantly evolving as seen due to the COVID-19 pandemic. COVID-19 is thus a perfect real-world use case for weakly supervised models.

\begin{figure*}[t]
\centering
  \includegraphics[width=0.8\textwidth]{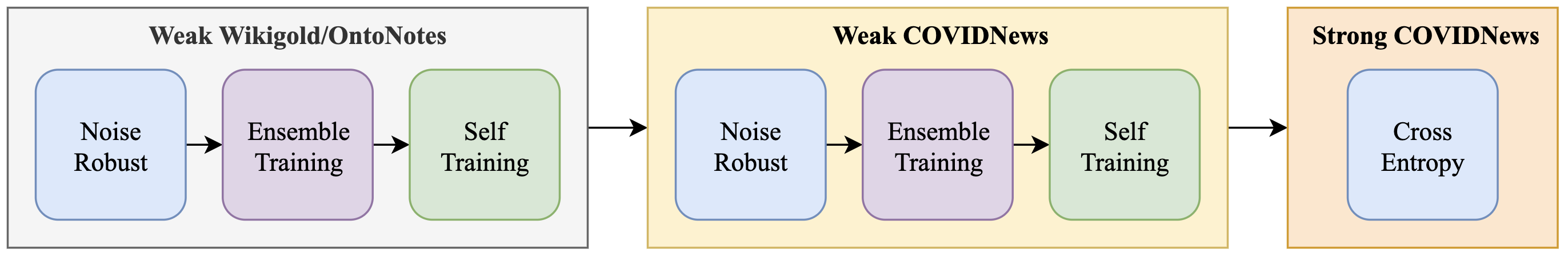}
  \caption{\model: a double-backboned weak-strong data finetuning architecture.}
  \label{fig:modelarchs}
\vspace{-1.3em}
\end{figure*}

However, weakly supervised methods are mostly tested on general-domain datasets rather than an emerging domain or topic. 
The inability of current state-of-the-art NER models to perform when given new biomedical topics such as COVID-19 preprints has been previously stated as a major gap in current NER applications \citep{arenotready}.
This was explained by a propensity for models to overfit to currently available training data and a lack of data in the target domain for models to learn such a complex emerging topic like COVID-19. We bridge this gap by proposing a domain-specific NER dataset called \dataset to evaluate these weakly supervised methods and providing suggestions of combining weak and strong data to address this issue (for a real example of weak and strong labels, see \Cref{fig:main_fig}). We expect that the data we publish will drive research around techniques to better adapt to new topics and the strong data we publish to unlock NER applications about COVID-19 and in the medical domain. We plan for our paper to specifically contribute to the relaunched and currently being improved ontology-based text mining tool BioCaster \citep{biocaster2021} for automatic monitoring and surveillance of disease outbreaks.

For our analysis of mixing weak and strong data, we build upon the recent weakly labelled NER model RoSTER \citep{Roster}. The model uses a noise robust loss function with noisy label removal, ensemble training and self training without the use of strong data to achieve best performance out of other distantly labelled methods. 
We propose \model, which improves upon RoSTER by performing cross-domain transfer learning over 3 training stages: The first two stages progressively train the model on out-of-domain and in-domain weak data; Afterwards, we finetune the model on in-domain strong data in the last stage (\Cref{fig:modelarchs}).

To summarise, this paper has the following contributions: \textbf{(1)} We propose a novel COVID-19 NER dataset with 13000 weakly labelled sentences generated by rule based methods and 3000 manually annotated sentences. To our knowledge, this is the first COVID-19 orientated NER dataset in English.
\textbf{(2)} We experiment with the data and provide insight into the effect from training with weak and strong data individually and when combined. \textbf{(3)} We propose \model, a cross-domain continual training framework, to best leverage strong data and multi-source weak data, and achieve state-of-the-art performance on \dataset.

%% file: AACL2022/002_dataset.tex
\section{Dataset: \dataset} \label{sec:dataset}


\vspace{-0.2em}
\paragraph{Data generation and filtering.}
The data consists of 13000 sentences gathered and weakly labelled using the BioCaster ontology-based text mining tool \citep{biocaster} with 3000 of the sentences also being manually annotated. BioCaster first generated the text for the dataset by scraping news articles from multiple local news providers and RSS feeds covering pandemic related topics between approximately January to August 2021. Once BioCaster collected sentences from its news sources, the entries were passed through a text classifer to further refine that selections were oriented towards disease outbreaks \citep{conway2009classifying}. The classifer selected was recently updated to use the pretrained PubMedBERT \citep{gu2021domain} as a backbone classifier with further finetuning on a binary document classification dataset made of alternating pandemic and normal type news. BioCaster generated part of the dataset from native English texts with additional entries translated from French, Indonesian and Mandarin to English using Language Weaver's Edge MT engine\footnote{\url{https://www.rws.com/language-weaver/edge/}}. The system finally filtered entries by removing entries from duplicate sources from the randomly sampled variety of articles chosen during selection. The weak labels of the dataset were then generated using BioCaster's rule-based method \citep{collier-etal-2010-ontology}. The method is made up of regular expression patterns in simple rule language (SRL), a tool built on top of DIAL \citep{dial}.

To ensure a high quality of final dataset entries, multiple filtering methods were implemented after this procedure to further prune text with errors. All candidate text was filtered out based on insufficient text lengths, non-ASCII characters involved and text duplicates. Additionally, texts were pruned based on number of grammatical mistakes per entry and finally through manual examination. Manually annotated strong data was labelled by a recent graduate working in the NLP domain. Challenging entries such as differentiating between virus and disease entities were flagged and resolved by discussion with a PhD student and Professor who served as experts in the biomedical NLP domain.\footnote{See \Cref{sec:data_filtering_details} for more details.}

\vspace{-0.6em}
\paragraph{Dataset entities and content.}
We employ 10 entity types: \texttt{Animal}, \texttt{Bacterium}, \texttt{Disease}, \texttt{Location}, \texttt{Organisation}, \texttt{Person},  \texttt{Product}, \texttt{Symptom}, \texttt{Time}, \texttt{Virus}.\footnote{See \Cref{sec:entity_defs} for definitions of the entity types.} The \texttt{Person} entity has been expanded to label human cases of a  disease as a group of people and the \texttt{Product} entity refers to manufactured articles in the medical domain used during the COVID-19 pandemic (eg. vaccines, face masks). The \texttt{Virus} entity is an especially useful common emerging label since the neighbouring text referencing COVID-19 changes whether it is a \texttt{Virus} or \texttt{Disease}, a common mislabel in weak data. In addition to this, the context developed as a result of the COVID-19 pandemic has produced emerging entities that current rule or knowledge-based labelling systems do not capture. These are however clear to the general public and human annotators which we demonstrate through examples found in \mbox{\dataset} in \mbox{\Cref{fig:additional_fig}}. Because the dataset has a heavy focus on the COVID-19 virus, new terminology is also featured surrounding vaccination, testing, variants, etc. Other viruses that gained exposure to the public due to the pandemic are also included in the dataset such as the Zika virus, MERS coronavirus and influenza virus. In general, emerging entities and unique text make \mbox{\dataset} tailored to pandemics and the medical domain while still providing some useful entities for general NER applications.

\vspace{-0.6em}
\paragraph{Inter-Annotator Agreement (IAA).}
To demonstrate the quality of the strong data when compared to the weak data we preform an inter-annotator agreement test. The method of Cohen's Kappa for inter-annotator agreement has been considered inaccurate for NER due to the task not having negative cases to fulfil the methods calculation \mbox{\citep{inter_ref}}. In our case, we recruited four additional validators with relevant background to re-annotate 100 randomly selected entries from our dataset. Annotators were given a comprehensive guideline on the labelling strategy and spent on average 90 minutes to read the guideline and complete the labels. We then computed pairwise F1 scores between each of the annotators and the original human annotated 100 strong labels from the dataset. We show this score along with the annotators score when compared to the weak data and the original strong data when compared to the weak data in \mbox{\Cref{tab:intertable}}. The human-labelled strong data has shown high agreement with the validators' labels, achieving >90\% F1 score, demonstrating that the human labelled strong data have high quality.

\begin{table}[htb]
\centering
\small
\begin{tabular}{lcc}
\toprule
\textbf{Test} & \textbf{F1} & \textbf{Std. Dev}\\
\midrule
Strong vs Weak & 46.2 & -  \\
Weak vs Validators & 49.8 & 2.35  \\
Strong vs Validators & 92.3 & 3.08 \\
\bottomrule
\end{tabular}
\caption{\label{tab:intertable} Dataset Inter-Annotator Agreement}
\vspace{-0.1cm}
\end{table}


\paragraph{Data statistics.}
\Cref{tab:datasetstats} summarises multiple metrics that describe the \dataset dataset. Of the 13000 weak data entries, 3000 of the same text have been manually annotated to provide the parallel strong labels with the equivalent 3000 weak labels also evaluated for fair comparison. We provide the total number of words, labelled words and entities for both strong and weak data. There is a noticeable difference in entity length with the average number of words in an entity being 1.489 and 1.890 for the respective 3000 weak and strong data entries. Longer entities are more challenging to fully label and explains how the weak labelling scheme tends to produce shorter labelled entities. This is especially noticable in types \texttt{Organisation} and \texttt{Bacterium} seen in \Cref{tab:weakdataperf} where weak data is evaluated directly against strong data. The weak data also had a lower number of entities per entry than the strong data which infers that the weak labelling scheme misses more ambiguous entities and that it is in general under labelled. We provide more analysis on weak vs. strong data in \Cref{sec:weak_against_strong} and data split generation in \Cref{sec:data_split}.

\begin{table}
\centering
\small
 \scalebox{0.92}{
\begin{tabular}{lrrr}
\toprule
\textbf{Metric} & \textbf{Weak} & \textbf{Weak-3k} & \textbf{Strong} \\
\midrule
Total Entries (Sentences) & 13000 & 3000 & 3000 \\
Total Words & 349913 & 80539 & 80539 \\
Total Labelled Words & 42692 & 9327 & 14786 \\
Total Entities & 28431 & 6263 & 7823 \\
Mean Entity Length & 1.50 & 1.50 & 1.89 \\
Percent Labelled Words & 12.2\% & 11.6\% & 18.4\% \\
Mean Entities Per Entry & 2.19 & 2.09 & 2.61 \\
\bottomrule
\end{tabular}
}
\caption{\label{tab:datasetstats} Generic statistics of \dataset}
\vspace{-0.4cm}
\end{table}





%% file: AACL2022/003_model.tex
\section{Model} \label{sec:model}

\vspace{-0.2em}
We build upon RoSTER \citep{Roster} which achieves the best performance among distantly-supervised methods. The model contains multiple stages to handle the weak data and its inherent noise. Starting with RoBERTa \citep{liu2019roberta} weights, the first step is noise robust training using generalised cross entropy ($\mathcal{L}_{\text{GCE}}$) with tunable parameters dictating noise robustness and noisy label removal. The second step uses ensemble training to improve model stability and the third step introduces contextualised augmentations and self-training with pre-trained RoBERTa embeddings.\footnote{See \Cref{sec:roster_details} for details of RoSTER.}


\vspace{-0.5em}
\paragraph{\model: \underline{Cont}inually-learned \underline{RoSTER}.} 
While RoSTER achieves strong performance on noisy data, it remains unclear what is the optimal strategy when both strong and weak data are presented. Additionally, in a real-world use case, we can also assume access to weak labels in other domains. We propose a continual learning approach called \model to adapt RoSTER for learning from out of domain weak data and in domain weak and strong data (\Cref{fig:modelarchs} represents the fine-tuning pipeline). The pipeline has three training stages: \textbf{(1)} We initially train a RoSTER model on out-of-domain weak data (grey box in \Cref{fig:modelarchs}). The out-of-domain data are from weak labels generated onto the Wikigold or OntoNotes dataset via knowledge bases (details explained in \Cref{sec:compared_models}). \textbf{(2)} Then we repeat RoSTER training on in-domain COVIDNews weak data (yellow box) and \textbf{(3)} finally finetuning on strong data with only the noise-robust loss. 


%% file: AACL2022/004_exp.tex
\section{Experiments}\label{sec:exp}

\vspace{-0.2em}
\paragraph{Compared models.}\label{sec:compared_models} We train four models and evaluate them on the \dataset test set. 
\textbf{(1)} We train the original RoSTER model with no backbone (initialised with original RoBERTa weights) on strong data. As the data are clean, the model is only trained with the noise robust loss with ensemble learning and self-training stages removed. 
\textbf{(2)} We train the model on 6000 lines of weak COVIDNews data (this creates a weak COVIDNews backbone) and then finetune on strong COVIDNews data the same as (1). \textbf{(3)} \& \textbf{(4)} In the double backbone approach (i.e., our full \model model), we first train RoSTER on either the weak labels from  the Wikigold dataset \citep{wikigold} or the OntoNotes5.0 dataset \citep{onto} followed by training on weak COVIDNews data and finally fine-tuning on strong COVIDNews data. This approach is visualised in \mbox{\Cref{fig:modelarchs}} in which \textbf{(1)} is made up of only the \texttt{Strong COVIDNews} box and \textbf{(2)} is made up of the \texttt{Weak COVIDNews} and \texttt{Strong COVIDNews} boxes. The Wikigold dataset contains 13041 lines of training data and 3 overlapping entity categories with COVIDNews out of 4 total entity categories. The OntoNotes5.0 dataset contains 59924 lines of training data and 5 overlapping entity categories with COVIDNews out of 18 total entity categories\footnote{Wikigold/OntoNotes weak data are from \citet{Roster}.}. 

\begin{figure}[htb]
    \centering
    \includegraphics[width=\linewidth]{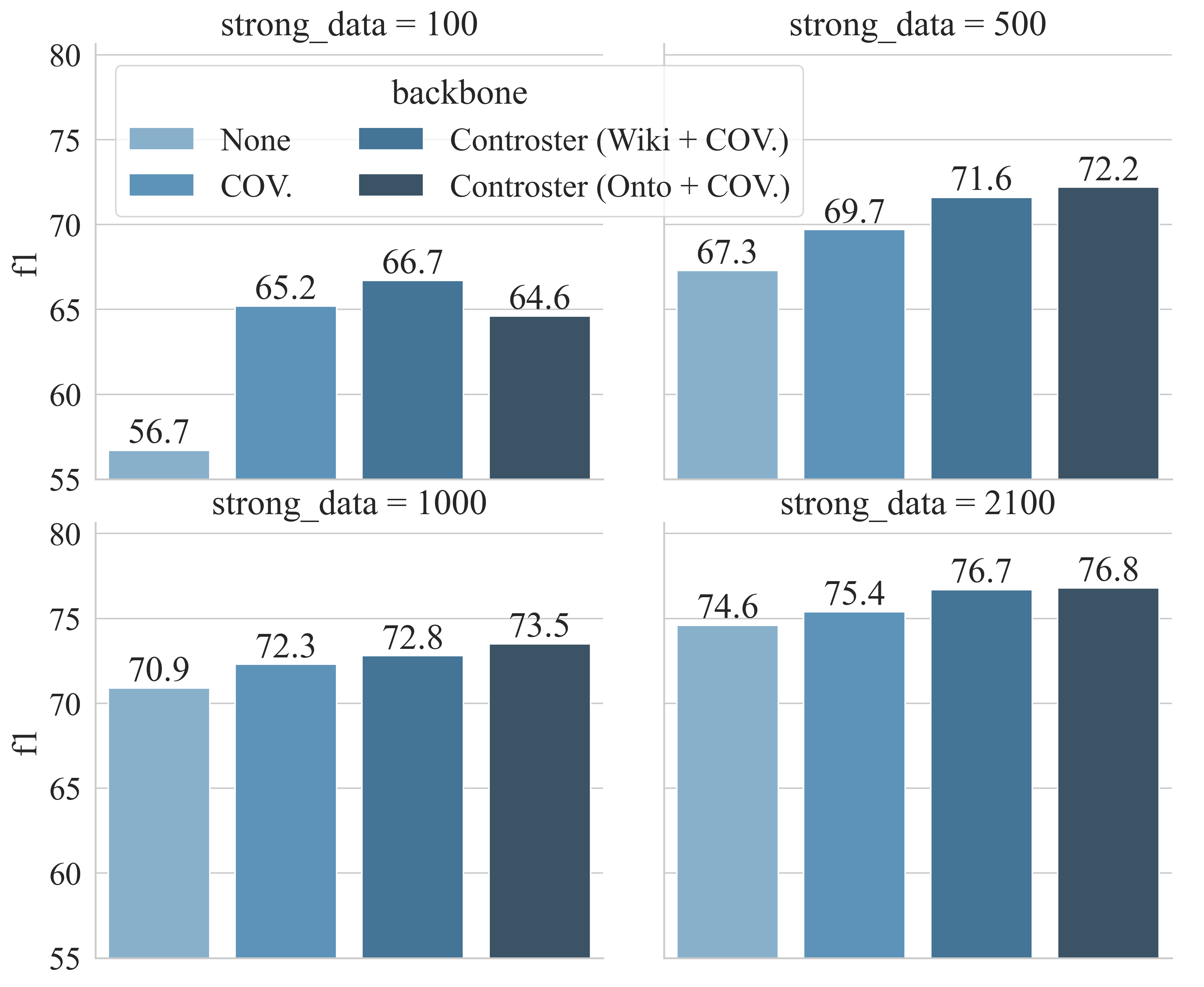}
    \caption{Main weak and strong data results. When using different number of strong data points, having the COVIDNews backbone (COV.) has always helped. Adding the Wiki/Onto backbone on top of COV. has also almost universally helped.}
    \label{fig:multidata}
    \vspace{-1.2em}
\end{figure}

\vspace{-0.5em}
\paragraph{Main results.}\label{sec:main_res}
\Cref{fig:multidata} shows the four model's F1-score performance.\footnote{Precision \& Recall follow the same trend (see \Cref{tab:multidata}).} The double backbone approach, i.e. \model, performs best for all quantities of strong data used for finetuning. Additionally, using a weak COVIDNews backbone performs noticeably better than the baseline without a backbone for all four amounts of finetuning strong data. The improvement in performance with using either a single or double backbone approach is greatest for experiments with 100 and 500 entries of strong data when compared to using 1000 or 2100. An example of this can be seen by looking at the improvement of 11.0 in F1 score (56.7 to 66.7) when the weak Wiki+COV. backbone was paired with 100 entries of strong data in comparison to an increase of 2.2 in F1 score (74.6 to 76.8) in the same scenario when 2100 entries of strong data were used. 

These findings support the use of combining weak data with strong data through the method of transfer learning for research and NER applications. We provide insight into the amounts of strong data necessary for effective combination of the two types of data.  We recommend using a weak data generated backbone in general NER models with the potential for profound impact in few-shot learning models that have a limited number of strong data. Similarly, in scenarios where only weak data is available we advise manually annotating a minimum of 100 sentences can lead to large improvements in NER model performance.

\begin{figure}[htb]
    \centering
        \vspace{-0.1cm}
    \includegraphics[width=0.8\linewidth]{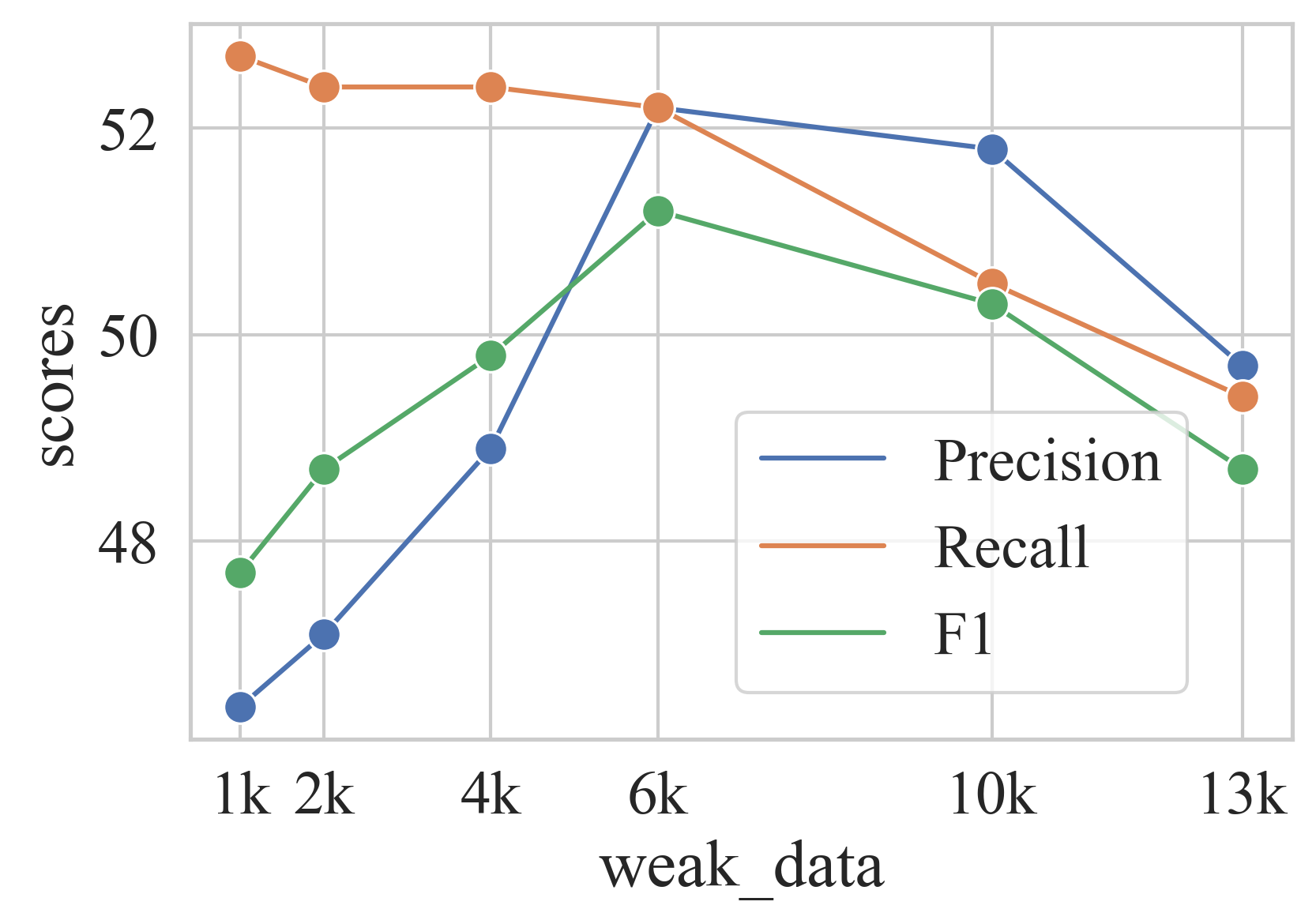}
    \vspace{-0.2em}
    \caption{Weak data performance of RoSTER.}
    \label{fig:indivdata}
    \vspace{-1.3em}
\end{figure}

\vspace{-0.5em}
\paragraph{Weak data study.}\label{sec:weak_data_study}
We investigate the magnitude of weak data for the models trained to be effective on the \dataset dataset. 
\Cref{fig:indivdata} shows the improvement in precision, recall and F1 scores as weak data scales with the RoSTER model. After surpassing the 6k entry mark, the weak data saturates and does not lead to improved performance for each additional entry. Alternatively, the strong data in \Cref{tab:multidata} demonstrates how the higher quality strong labels continue to scale with additional data. Varying amounts of weak data were also evaluated for the COVIDNews backbone prior to finetuning on strong data. \Cref{tab:combineddata} in Appendix shows tests with 2000, 6000 and 10000 weak entries paired with 100, 500, 1000 and 2100 strong entries. Although using a weak data backbone was clearly beneficial, the amount of weak data to train the backbone was fairly insignificant.


%% file: AACL2022/005_discussion.tex
\section{Further Discussion} \label{sec:furthur_discussion}
\vspace{-0.3em}
\paragraph{Combining weak \& strong data.}
We believe a wide and deep strategy should be used when combining weak \& strong data. This means training on weak data to embed the model with an expansive breadth of knowledge over all entities and then fine-tuning on strong data to overwrite noise in the weak data and generate more sophisticated ability in NER. We suggest two avenues for further research to maximise utility of weak data when paired with strong data. An improvement in noise reduction techniques via improved loss functions or model architecture will serve beneficial in allowing model performance to continue to scale with increases in weak data. Additionally, investigation into how weak data generated from specific rules saturates individually and after fine-tuning with strong data will also lead to improved knowledge on mixing the different forms of labels.

\vspace{-0.6em}
\paragraph{Out-of-domain weak data.}
We observed that using the two stage backbone pipeline of first training on a different NER dataset and then following through with training on \dataset weak and strong data led to a further improvement in performance across varying \dataset weak and strong data lengths. Even though the Wikigold dataset had only three overlapping entity categories with \dataset and OntoNotes5.0, the performance still improved in part due to the overlapping categories being the prominent ones in \dataset (eg. organisation, person, location, date). The additional variety of labels in the crossover entities led to improved precision, recall and F1 scores in those categories and overall scores. We implore future works to evaluate the impact of overlapping and non-overlapping categories from cross-domain weak data backbones while considering overlapping category definition similarities and differences. Overall, similar to how \citet{multibert} determined the importance of using multiple sources for text to be distantly labelled, we conclude it is also beneficial to use different weak labelling techniques to create a diverse collection of weak data. We implore future work to investigate the diversity necessary for optimal combination of weak data from different sources.

In Appendix, we include a dedicated related work section (\Cref{sec:RW}) for interested readers.

\vspace{-0.1em}

%% file: AACL2022/006_conclusion.tex
\section{Conclusion}
\vspace{-0.3em}
We presented \dataset, an English COVID-19 Named Entity Recognition dataset in the pandemic news domain, addressing current NER models' lack of ability to tackle new and out-of-domain topics. 
We labelled 13000 entries using a rule-based system to generate weak labels and 3000 entries using hand annotation to generate strong labels. 
We further proposed a continual learning approach called \model that transfers knowledge learned in both out-of-domain and in-domain weak data. After finetuning on strong in-domain data, \model achieved state-of-the-art performance on our proposed dataset. We further provide detailed and thorough analysis into how to successfully combine both types of data and suggest promising avenues for future research. We think that the dataset we provide and the findings we conclude will be beneficial to other NER applications, such as improving the evaluation and ability of the BioCaster pandemic surveilling tool. We hope that our work drives more research in leveraging a combination of weak and strong data to improve performance on new topics such as the COVID-19 pandemic.

%% file: AACL2022/007_acknowledgments.tex
\section*{Acknowledgements}

We are grateful to RWS Language Weaver for use of their neural MT engine. We also thank Qianchu Liu, Parth Shah, Chandni Bhatt and Marko Popovic for contributing to the inter-annotator agreement.

%% file: AACL2022/008_appendix.tex
\appendix


\section{More Dataset Details}\label{sec:data_details}

\subsection{Data Generation and Filtering Details}\label{sec:data_filtering_details}

The text used in the dataset and the corresponding manually annotated strong labels had some deviation in metrics depending on the original language translated from. Additional statistics of the dataset for each text language origin can be seen in \Cref{tab:extralanguagemetrics}. The metrics suggest that Mandarin was an especially useful and unique language to generate data from as it introduced many longer entities due to person titles and location addresses being more verbose. Other languages like French introduced text with fewer entities per sentence than others.

\begin{table}[htb]
\centering
\small
 \scalebox{0.8}{
\begin{tabular}{lrrrrr}
\toprule
\textbf{Metric} & \textbf{Total} & \textbf{Eng.} & \textbf{Fre.} & \textbf{Ind.} & \textbf{Man.} \\
\midrule
Total Entries & 3000 & 1500 & 505 & 500 & 495 \\
Mean Entity Length & 1.89 & 1.79 & 1.80 & 1.68 & 2.46 \\
Percent Labelled Words & 18.4\% & 17.5\% & 15.4\% & 18.9\% & 22.9\% \\
Mean Entities Per Entry & 2.61 & 2.71 & 2.09 & 2.70 & 2.74 \\
\bottomrule
\end{tabular}
}
\caption{\label{tab:extralanguagemetrics} Generic statistics of \dataset strong data separated by language}
\end{table}

For filtering, texts that were less than 4 words, less than 15 characters or greater than 500 characters were removed. Duplicate sentences were also filtered out and poorly structured entries were identified using the LanguageTool grammar checking API\footnote{\url{https://languagetool.org/}}. This checked and removed entries with grammar, punctuation and syntactical mistakes.

\subsection{Entity Definitions}\label{sec:entity_defs}

The exact definitions of the 10 entity types included in \dataset can be found in \Cref{tab:entitydefs}.

\begin{table*}
\small
\centering
\begin{tabular}{rl}
\toprule
\textbf{Entity Type} & \textbf{Definition} \\
\midrule
\texttt{Animal} & Multi-cell organisms that are eukaryotes of the kingdom Animalia, other than humans. \\	
\texttt{Bacterium} & Single-celled prokaryotic microorganisms of the bacteria domain.  \\	
\texttt{Disease} & A disorder of a structure or function that affects an organism, associated with specific phenotypes. \\		
\texttt{Location} & A politically or geographically defined location for example a region, a province, a town. \\	
\texttt{Organisation} & Named corporate, governmental, or other organisational entity. \\	
\texttt{Person} & A person or group of persons. \\		
\texttt{Product} & Medical articles or substances manufactured and used throughout pandemics.   \\		
\texttt{Symptom} & Phenotypic descriptions of any abnormal morphology, physiology or behaviour.  \\	
\texttt{Time} & Temporal expressions that can be anchored on a timeline. \\	
\texttt{Virus} & A disease causing infectious agent that is non-living.  \\		
\bottomrule
\end{tabular}
\caption{\label{tab:entitydefs} Entity Type Definitions}
\vspace{-0.2cm}
\end{table*}

\subsection{Weak vs. Strong Data}\label{sec:weak_against_strong}

\begin{table}[t]
\centering
\small
\begin{tabular}{lcc}
\toprule
\textbf{Entity Type} & Strong & Weak \\
\midrule
\texttt{Animal} & 177 & 201 \\		
\texttt{Bacterium} & 25 & 12 \\		
\texttt{Disease} & 641 & 612 \\		
\texttt{Location} & 1703 & 1568 \\		
\texttt{Organisation} & 1076 & 270\\		
\texttt{Person} & 2652 & 2370 \\		
\texttt{Product} & 233 & 203 \\		
\texttt{Symptom} & 121 & 146 \\		
\texttt{Time} & 799 & 697 \\		
\texttt{Virus} & 396 & 184 \\		
\midrule
Total & 7823 & 6263 \\
\bottomrule
\end{tabular}
\caption{Entity counts in \dataset}
\vspace{-0.7cm}
\label{tab:entity_stats}
\end{table}

\begin{figure*}[t]
    \centering
    \includegraphics[width=\linewidth]{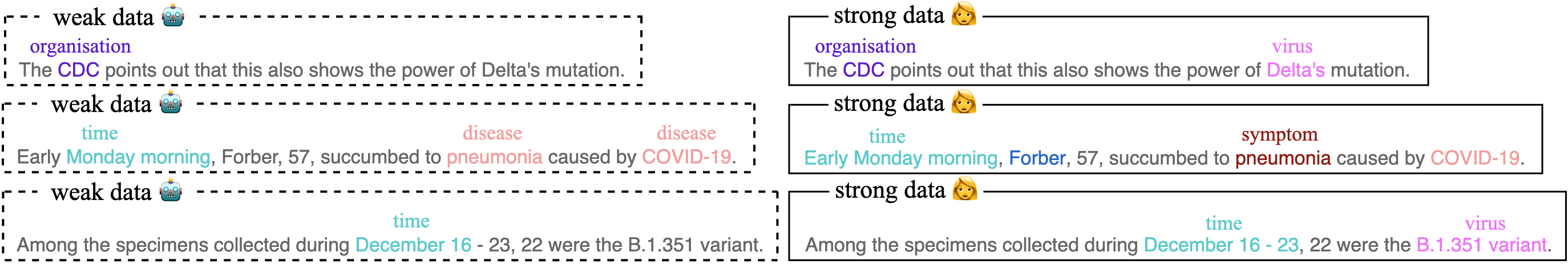}
    \caption{Additional examples from the \dataset dataset}
    \label{fig:additional_fig}
    \vspace{-0.2cm}
\end{figure*}

We further investigate the differences in performance and style between the rule based weak data generation method and human annotated strong data. 
\Cref{tab:entity_stats} shows the difference in entity count for each category between the two types of data. \texttt{Organisation} is noticeably out numbered in the strong data case which can be explained by the category requiring more in depth understanding of contextual knowledge as that can change it being classified as a \texttt{Location} or \texttt{Organisation} (eg. ``The White House'').
Another notable difference is that larger groups of words are categorised as entities in the strong labels when compared to the weak. The strong data contains 2.5 times more entities containing greater than three words due to the difficulty in labelling longer entities. 
The \texttt{Symptom}, \texttt{Disease} and \texttt{Virus} entity categories in the dataset significantly orient the dataset towards the COVID-19 pandemic. The difference between the three categories are challenging to distinguish, examples of which are shown in \Cref{fig:additional_fig}. \Cref{tab:weakdataperf} shows a detailed breakdown of the performance of weak data when evaluated directly against strong data and \Cref{tab:langaugeperf} shows the performance across the different languages the text was translated from.

\begin{table}[t]
\centering
\small
\begin{tabular}{llllr}
\toprule
\textbf{Entity Type} & \textbf{Pre.} & \textbf{Rec.} & \textbf{F1} & \textbf{Support} \\
\midrule
\texttt{Animal} & 62.2 & 70.6 & 66.1 & 177 \\		
\texttt{Bacterium} & 33.3 & 16.0 & 21.6 & 25 \\		
\texttt{Disease} & 66.2 & 63.2 & 64.6 & 641 \\		
\texttt{Location} & 57.0 & 52.4 & 54.6 & 1703 \\		
\texttt{Organisation} & 33.3 & 8.4 & 13.4 & 1076 \\		
\texttt{Person} & 46.7 & 41.7 & 44.0 & 2652 \\		
\texttt{Product} & 63.1 & 54.9 & 58.7 & 233 \\		
\texttt{Symptom} & 46.6 & 56.2 & 50.9 & 121 \\		
\texttt{Time} & 68.4 & 59.7 & 63.8 & 799 \\		
\texttt{Virus} & 49.5 & 23.0 & 31.4 & 396 \\		
\midrule
Weighted Avg & 51.8 & 43.3 & 46.2 & 7823 \\
\bottomrule
\end{tabular}
\caption{\label{tab:weakdataperf} Weak Data Performance}
\vspace{-0.2cm}
\end{table}

\begin{table}[t]
\centering
\small
\begin{tabular}{llllr}
\toprule
\textbf{Entry Language} & \textbf{Entries} & \textbf{Pre.} & \textbf{Rec.} & \textbf{F1} \\
\midrule
Combined & 3000 & 59.4 & 44.3 & 49.7  \\
English & 1500 & 60.3 & 41.0 & 47.5 \\
French & 505 & 62.3 & 53.3 & 56.5 \\
Indonesian & 500 & 62.6 & 49.4 & 54.1 \\
Mandarin & 495 & 53.9 & 42.6 & 46.1 \\
\bottomrule
\end{tabular}
\caption{\label{tab:langaugeperf} Weak Data Performance Across Languages}
\vspace{-0.5cm}
\end{table}



\subsection{Data Split}\label{sec:data_split}
To best split the data into training, validation and test sets, a unique Monte Carlo technique was implemented to insure entities with limited labels such as \texttt{Bacterium} were allocated in proper amounts to each partition. An optimal distribution of all entities was generated and 10000 random iterations of the input data was split and scored based on fractional proximity to the optimal distribution. The split dataset partitions had 2100/300/600 entries in train, validation and test sets respectively.

\section{RoSTER Details}\label{sec:roster_details}

\vspace{-0.5cm}

Here we explain the RoSTER methodology in greater detail. The first stage is known as the noise-robust learning stage and introduces two hyperparameters for adjusting to noisy labels. \citep{Roster} uncovers that cross entropy loss is useful for model convergence but is sensitive to noise while mean absolute error (MAE) loss is robust to noise at the cost of convergence. The generalised cross entropy loss uses a \emph{q} parameter to adjust cross entropy towards CE loss by lowering and towards MAE loss by raising. A thersholding parameter is introduced to remove incorrect labels during the training process. The parameter $\uptau$ is used as a threshold for comparing model predictions with distant labels. If there are differences between model predictions and distant labels greater than the threshold, the model omits those labels when updating weights.
\vspace{0.5cm}

\begin{equation}
\mathcal{L}_{\text{GCE}} = \sum_{i=1}^{n} w_{i}\frac{1-f_{i,y_{i}}(x;\theta)^{q}}{q}
\end{equation}

RoSTER also implements ensemble and self training stages to improve results on distantly labelled data. The ensemble stage uses a \emph{K} parameter to determine the number of models trained using different seeds and a final model is employed to approximate the performance of trained models by minimising Kullback–Leibler (KL) divergence loss. Prior to self-training, contextualised augmentations are generated using PLM's like RoBERTa. Then the model trains on an unlabelled version of the corpus to leverage knowledge embedded in the selected PLM while generalising model predictions to tokens removed by noisy label removal. Self-training is done by polarising predictions during iterations by squaring high-confidence predictions and normalising low-confidence predictions.

\section{More Experimental Details}\label{sec:more_exp_details}

\paragraph{Weak-strong main results full table (\Cref{tab:multidata}).} In the main text we showed performance of \model and its ablated versions' in \Cref{fig:multidata}. Here, we provide a more detailed view of the same data, listing also Precision and Recall scores in \Cref{tab:multidata}.

\begin{table}[htb]
\centering
\small
\begin{tabular}{lclll}
\toprule
\textbf{W. Backbone} & \textbf{S. Tuning} & \textbf{Pre.} & \textbf{Rec.} & \textbf{F1} \\
\midrule

None & 100 & 50.9 & 65.2 & 56.7 \\
COV. & 100 & 59.9 & 71.7 & 65.2 \\
Wiki. + COV. & 100 & \textbf{62.1} & \textbf{72.5} & \textbf{66.7} \\
Onto. + COV. & 100 & 60.2 & 69.9 & 64.6 \\
\midrule
None & 500 & 62.8 & 73.0 & 67.3 \\
COV. & 500 & 66.3 & 73.9 & 69.7 \\
Wiki. + COV. & 500 & 68.0 & 75.7 & 71.6 \\
Onto. + COV. & 500 & \textbf{68.5} & \textbf{76.7} & \textbf{72.2} \\
\midrule
None & 1000 & 66.7 & 75.8 & 70.9 \\
COV. & 1000 & 69.1 & 76.1 & 72.3 \\
Wiki. + COV. & 1000 & 69.6 & 76.6 & 72.8 \\
Onto. + COV. & 1000 & \textbf{70.2} & \textbf{77.3} & \textbf{73.5} \\
\midrule
None & 2100 & 71.7 & 77.9 & 74.6 \\
COV. & 2100 & 72.9 & 78.2 & 75.4 \\
Wiki. + COV. & 2100 & 73.9 & \textbf{79.9} & 76.7 \\
Onto. + COV. & 2100 & \textbf{74.2} & 79.7 & \textbf{76.8} \\

\bottomrule
\end{tabular}
\caption{\label{tab:multidata} Main weak and strong data results
}
\vspace{-0.5cm}
\end{table}

\paragraph{Weak backbone saturation data (\Cref{tab:indivdata}).} Since we have an in-domain rule-based weak labeller, why not generate as much in-domain weak data as possible? As mentioned in the main text \Cref{fig:indivdata}, we found that in-domain weak data only helps up to a certain point. Here we list the exact precision, recall and F1 results used for plotting the figure of reference in the main text (\Cref{tab:indivdata}).

\begin{table}[htb]
\centering
\small
\begin{tabular}{llll}
\toprule
\textbf{Weak Data} & \textbf{Pre.} & \textbf{Rec.} & \textbf{F1} \\
\midrule
1000 entries & 46.4 & \textbf{52.7} & 47.7 \\
2000 entries & 47.1 & 52.4 & 48.7 \\
4000 entries & 48.9 & 52.4 & 49.8 \\
6000 entries & \textbf{52.2} & 52.2 & \textbf{51.2} \\
10000 entries & 51.8 & 50.5 & 50.3 \\
13000 entries & 49.7 & 49.4 & 48.7 \\
\bottomrule
\end{tabular}
\caption{\label{tab:indivdata} Performance of RoSTER when varying number of weak data }
\vspace{-0.5cm}
\end{table}

\paragraph{Weak data study (\Cref{sec:weak_data_study}) full table (\Cref{tab:combineddata}).} In the main text we discussed varying amounts of weak data when pretraining on \dataset. Here we attach the full table (\Cref{tab:combineddata}) for reference.

\begin{table}[htb]
\centering
\small
\begin{tabular}{lclll}
\toprule
\textbf{W. Backbone} & \textbf{S. Tuning} & \textbf{Pre.} & \textbf{Rec.} & \textbf{F1} \\
\midrule
None & 100 & 50.9 & 65.2 & 56.7 \\
Weak 2000 & 100 &  59.9 & 69.2 & 64.1 \\
Weak 6000 & 100 & 60.4 & 70.6 & 65.0 \\
Weak 10000 & 100 & \textbf{61.5} & \textbf{70.9} & \textbf{65.7} \\
\midrule
None & 500 & 62.8 & 73.0 & 67.3 \\
Weak 2000 & 500 & \textbf{67.1} & 74.2 & \textbf{70.3} \\
Weak 6000 & 500 & 66.3 & 73.9 & 69.7 \\
Weak 10000 & 500 & 66.9 & \textbf{74.3} & 70.2 \\

\midrule
None & 1000 & 66.7 & 75.8 & 70.9 \\
Weak 2000 & 1000 & 69.6 & 76.1 & 72.6 \\
Weak 6000 & 1000 & 69.1 & 76.1 & 72.3 \\
Weak 10000 & 1000 & \textbf{70.2} & \textbf{76.7} & \textbf{73.2} \\

\midrule
None & 2100 & 71.7 & 77.9 & 74.6 \\
Weak 2000 & 2100 & 72.7 & 77.2 & 74.7 \\
Weak 6000 & 2100 & \textbf{72.9} & \textbf{78.2} & \textbf{75.4} \\
Weak 10000 & 2100 & 72.6 & 77.5 & 74.9 \\

\bottomrule
\end{tabular}
\caption{\label{tab:combineddata} Weak data quantities with strong data results
}
\vspace{-0.5cm}
\end{table}

\section{Related Work}\label{sec:RW}

Our work is related to other COVID-19 datasets in the NER domain. \citet{vietcovid} introduced a COVID-19 NER dataset for the low resource language of Vietnamese and \citet{italiancovid} provided a NER dataset based on medical records in Italian. \citet{englishcovid} scraped and annotated COVID-19 related tweets, generating a knowledge base but labelling events (eg. tested positive, can not test) as opposed to entities necessary for NER. Our dataset provides the first COVID-19 NER dataset in English with distantly supervised weak data and human annotated strong data.

Our analysis of combining weak and strong data is related to previous methods which successfully utilise either types of data to improve performance. \citet{bond2020} implemented the use of pre-trained language models with subsequent self-training with weak labels generated through knowledge bases to improve model performance. \citet{Jiang2021} architected a multistage pipeline involving pre-training on unlabelled data, weak label completion, a noise robust loss function and fine tuning on strong data to effectively null the impact of noise. We build upon these works and provide insight into using the two forms of data together in addition to using cross-domain datasets on an emerging topic such as COVID-19.